\documentclass{article}
\usepackage[numbers]{natbib}

\usepackage{arxiv}

\usepackage[utf8]{inputenc} 
\usepackage[T1]{fontenc}    
\usepackage{hyperref}       
\usepackage{url}            
\usepackage{booktabs}       
\usepackage{amsfonts}       
\usepackage{amsmath}        
\usepackage{nicefrac}       
\usepackage{microtype}      
\usepackage{lipsum}         
\usepackage{graphicx}
\usepackage{natbib}
\usepackage{doi}
\usepackage{graphicx}
\usepackage{subcaption}

\title{
Personalized Image Generation from an Author’s Writing Style
}

\author{ 
	{\hspace{1mm}$^\ast$Sagar Gandhi} \\
	Joyspace AI\\
	\texttt{sagar@joyspace.ai} \\
	\And
	{\hspace{1mm}$^\ast$Vishal Gandhi} \\
	Joyspace AI\\
	\texttt{vishal@joyspace.ai} \\
}

\renewcommand{\headeright}{} 
\renewcommand{\undertitle}{}
\renewcommand{\shorttitle}{}

\begin{document}
\maketitle
\pagestyle{plain} 

\renewcommand*{\thefootnote}{\fnsymbol{footnote}}
\footnotetext[1]{These authors contributed equally to this work.}

\begin{abstract}
Translating nuanced, textually-defined authorial writing styles into compelling visual representations presents a novel challenge in generative AI. This paper introduces a pipeline that leverages Author Writing Sheets (AWS) - structured summaries of an author's literary characteristics - as input to a Large Language Model (LLM, Claude 3.7 Sonnet). The LLM interprets the AWS to generate three distinct, descriptive text-to-image prompts, which are then rendered by a diffusion model (Stable Diffusion 3.5 Medium). We evaluated our approach using 49 author styles from Reddit data, with human evaluators assessing the stylistic match and visual distinctiveness of the generated images. Results indicate a good perceived alignment between the generated visuals and the textual authorial profiles (mean style match: $4.08/5$), with images rated as moderately distinctive. Qualitative analysis further highlighted the pipeline's ability to capture mood and atmosphere, while also identifying challenges in representing highly abstract narrative elements. This work contributes a novel end-to-end methodology for visual authorial style personalization and provides an initial empirical validation, opening avenues for applications in creative assistance and cross-modal understanding.
\end{abstract}


\section{Introduction}
\label{sec:introduction}

Authorial style, the unique fingerprint an author leaves on their work through choices in plot, characterization, language, and thematic development, is crucial for creating distinctive and engaging narratives. Personalizing generative AI to capture such individual styles has significant implications for interactive storytelling, creative writing assistance, and educational applications. Recent work by \textit{Lan et al.} \cite{Kumar_StoryStyle_2025} introduced the concept of ``Author Writing Sheets'' (AWS), a structured representation of an author's narrative characteristics, demonstrating their effectiveness in personalizing textual story generation. However, the potential to translate these rich, textually-defined authorial styles into the visual domain remains largely unexplored. The rapid advancements in text-to-image generation models, such as diffusion models \cite{Rombach_LatentDiffusion_2022}, offer new avenues for nuanced visual content creation, presenting an opportunity to bridge this gap and explore how an author's literary essence can be visually embodied.

This paper investigates the translation of nuanced, textually-defined authorial writing styles, as captured by Author Writing Sheets, into compelling visual representations. We propose and evaluate a pipeline that leverages a Large Language Model (LLM) to interpret these sheets and generate descriptive prompts for a text-to-image diffusion model. Our primary research questions are:
\begin{enumerate}
    \item[\textbf{RQ1:}] To what extent can abstract literary characteristics from an AWS be effectively translated into coherent and representative visual imagery?
    \item[\textbf{RQ2:}] What is the role of the intermediary LLM in interpreting the AWS and generating effective visual summary prompts?
    \item[\textbf{RQ3:}] Can the generated images be considered visually personalized to the author's style, distinguishing them from generic artwork?
    \item[\textbf{RQ4:}] How can the faithfulness and stylistic consistency of such cross-modal translations be systematically evaluated?
\end{enumerate}

To address these questions, we introduce a novel pipeline that first parses AWS data derived from Reddit authors (as presented in \cite{Kumar_StoryStyle_2025}). Second, a Large Language Model (Claude 3.7 Sonnet) interprets the cleaned AWS text to generate three distinct, descriptive text-to-image prompts designed to capture the author's aesthetic, mood, and characteristic themes. Finally, Stable Diffusion 3.5 Medium generates images based on these prompts. Our contributions are: (1) a novel pipeline for visual authorial style personalization, translating structured textual summaries of writing style into sets of stylistically cohesive images; (2) a methodology for using LLMs to generate targeted, stylistically-aware image prompts from these summaries; and (3) an empirical human evaluation of the generated images, focusing on their perceived stylistic consistency with the author's textual profile and their visual distinctiveness. The generated images, prompts, and author sheet identifiers are made available.\footnote{The dataset can be found at: [Link to your GitHub/Zenodo repository - to be added]}

The remainder of this paper is structured as follows: Section~\ref{sec:related_work} reviews related work. Section~\ref{sec:methodology} details our pipeline. Section~\ref{sec:experimental_setup} describes the experimental setup and evaluation design. Section~\ref{sec:results} presents our findings, followed by a discussion in Section~\ref{sec:discussion}. Finally, Section~\ref{sec:conclusion} concludes the paper.

\section{Related Work}
\label{sec:related_work}

Our work bridges personalization in generative AI with cross-modal style translation, specifically from textual authorial characteristics to visual representations.

\subsection{Personalization in Generative AI}
Personalizing generative models to reflect individual user characteristics or styles is a growing area of research.
The foundational work for our study by \textit{Kumar et al.} \cite{Kumar_StoryStyle_2025} focuses on textual personalization in story generation. They introduced the ``Author Writing Sheet'' (AWS), a structured summary of an author's implicit story-writing characteristics across narrative dimensions like Plot, Creativity, Development, and Language Use. This AWS, derived by comparing an author's stories to typical LLM-generated responses, is then used to guide an LLM to generate new stories in the author's style using tailored persona descriptions and story rules.
Prior research has also explored personalizing LLM outputs for various tasks, though often with shorter-form text. For instance, personalization has been applied to dialogue generation \cite{zhang_personalization_survey_2024}, and efforts have been made in tailoring news outputs \cite{krubinski_headline_generation_2024} and product recommendations or reviews \cite{li_prompt_learning_2023}. Many of these approaches infer user traits or preferences from historical data to tailor outputs \cite{wang_user_study_2024, li_review_generation_2019}. Our work differs by focusing on long-form creative generation and translating a deeply characterized authorial style into a different modality (visual) rather than generating text in a similar style. Furthermore, unlike role-playing approaches that often use predefined, well-known personas \cite{chen_ai_theater_2024}, we leverage inferred personas from an author's unique writing history.

\subsection{Text-to-Image Generation}
The generation of images from textual descriptions has seen remarkable progress, evolving from early Generative Adversarial Networks (GANs) \cite{goodfellow_gan_2014_placeholder} to the current prominence of diffusion models \cite{sohl_dickstein_diffusion_2015_placeholder, ho_denoising_diffusion_2020_placeholder}. Models like DALL-E \cite{openai_dalle_2021_placeholder}, Imagen \cite{saharia_imagen_2022_placeholder}, and Stable Diffusion \cite{Rombach_LatentDiffusion_2022} can now produce high-fidelity, coherent images from complex textual prompts.
A key challenge in text-to-image generation is achieving fine-grained control over the output, particularly concerning artistic style, aesthetics, and specific visual attributes. While prompt engineering is a common technique, more sophisticated methods have emerged. These include incorporating explicit stylistic prefixes or modifiers in prompts \cite{radford_clip_2021_placeholder}, fine-tuning models on specific artistic styles or subjects such as with DreamBooth \cite{Ruiz_DreamBooth_2023}, using image embeddings as style conditions through techniques like Textual Inversion \cite{Gal_TextualInversion_2022}, and employing auxiliary control networks like ControlNet \cite{zhang_controlnet_2023} to guide generation based on structural or semantic inputs. Our work leverages detailed textual descriptions of an author's \textbf{literary} style, aiming to translate these abstract narrative characteristics into corresponding visual aesthetics, a nuanced form of stylistic control.

\subsection{Cross-Modal Style Transfer and Representation}
Translating style or semantic content across different modalities is a complex but increasingly explored area. While text-to-image generation is inherently cross-modal, our focus is specifically on transferring a \textbf{textually-defined authorial style} to visual outputs.
Classic neural style transfer \cite{gatys_nst_2016_placeholder} demonstrated transferring the artistic style from one image to another. More recent efforts have explored translating textual descriptions of mood or genre into musical compositions \cite{musiclm_placeholder_2023} or generating textual narratives with stylistic considerations from visual inputs \cite{chen_multimedia_tang_2024}. The core challenge lies in learning a disentangled representation of style that can be mapped and applied across modalities. Our research contributes to this area by investigating whether the stylistic signals present in an author's literary works, as summarized in an AWS, can be effectively decoded and re-encoded into a visual style by a text-to-image model, mediated by an LLM interpreter. This moves beyond generic text-to-image generation by attempting to instill a specific, deeply characterized authorial persona into the visual output.

\section{Methodology: Visualizing Authorial Styles}
\label{sec:methodology}

\begin{figure}[ht]
\centering
\includegraphics[width=\columnwidth]{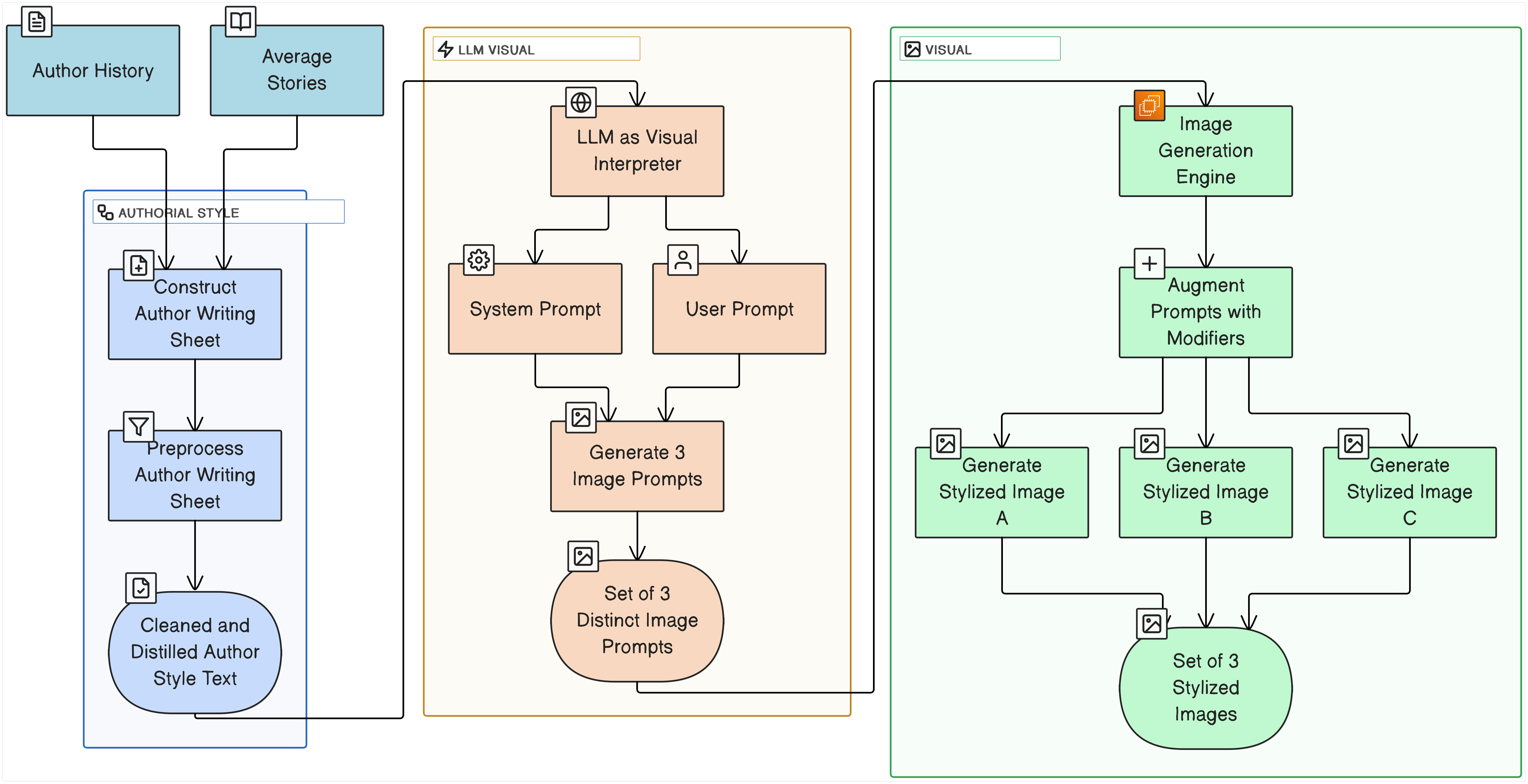} 
\caption{Overview of the pipeline for translating Author Writing Sheets into visual representations.}
\label{fig:pipeline_overview}
\end{figure}

To translate textual authorial styles into visual representations, we developed a multi-stage pipeline, as illustrated in Figure~\ref{fig:pipeline_overview}. The process begins with an Author Writing Sheet (AWS) as input, which is then interpreted by a Large Language Model (LLM) to generate a set of descriptive text-to-image prompts. These prompts subsequently guide a diffusion model to produce three distinct images intended to visually embody the author's unique style.

\subsection{Author Writing Sheets (AWS) as Stylistic Input}
\label{ssec:aws_input}
The foundation of our approach is the Author Writing Sheet (AWS), a structured, multi-faceted textual representation of an author's distinctive writing characteristics, as conceptualized by \textit{Lan et al.} \cite{Kumar_StoryStyle_2025}. Each AWS, derived from an analysis of an author's past works (specifically Reddit authors in our case), distills their tendencies across narrative categories such as Plot, Creativity, Development, and Language Use. This rich textual summary serves as the primary input for our visual style translation pipeline.

To effectively leverage the AWS for visual prompt generation, a preprocessing stage is employed. The raw AWS, typically stored in JSON format and containing structural tags (e.g., \textit{combined-author-sheet}) and metadata (e.g. evidence annotations, timestamps), is refined into a coherent narrative block. This involves extracting the core stylistic assertions and removing non-stylistic artifacts and formatting elements, thereby producing a distilled representation of the author's style optimized for comprehension by a Large Language Model. The objective is to provide the LLM with a clean, concentrated textual essence of the author's unique literary fingerprint.

\subsection{LLM-driven Cross-Modal Translation to Visual Prompts}
\label{ssec:llm_prompt_gen}
The core innovation in our methodology lies in utilizing a Large Language Model (LLM) – specifically Claude 3.7 Sonnet – as a sophisticated interpreter to bridge the gap between the abstract, textually-defined authorial style and the concrete, descriptive language required for text-to-image generation. This approach was chosen over potentially more brittle rule-based systems or data-intensive fine-tuning strategies, given the LLM's inherent capabilities in nuanced text understanding, abstract reasoning, and structured creative generation. The LLM's role is to translate the multifaceted literary style into actionable visual directives.

This translation is guided by a carefully designed prompting strategy. The LLM is primed with a system prompt that frames its task as that of an expert visual semiotician, tasked with identifying visual correlates for textual stylistic cues. For instance, textual descriptions of tone might map to visual mood or color palettes, recurring literary themes to subject matter, and narrative complexity to artistic style or composition. A critical aspect of this guidance is to encourage the LLM to synthesize the \textbf{overall aesthetic essence} and recurring patterns, rather than attempting a literal, and often impossible, visualization of every specific textual claim (e.g., abstract plot mechanics).

The preprocessed AWS text is then provided to the LLM, which is instructed to generate three distinct text-to-image prompts. This number was chosen to explore varied visual facets of a single authorial style. Each prompt is formulated as a rich, comma-separated list of visual descriptors encompassing main subject, mood, artistic style, lighting, and other technical aspects designed for optimal interpretation by diffusion models. The LLM's output is a structured JSON object containing these three targeted prompts, effectively converting the author's literary DNA into a set of visual blueprints.

\subsection{Image Generation with Stable Diffusion}
\label{ssec:image_generation}
The three descriptive prompts generated by Claude serve as input to Stable Diffusion 3.5 Medium, a latent diffusion model \cite{Rombach_LatentDiffusion_2022}, for the final image synthesis step. Each prompt is processed independently, resulting in three distinct images per author style sheet.
To enhance visual quality, standard positive prompt modifiers such as ``8k, highly detailed, masterpiece, perfect composition, intricate details, professional quality, cinematic lighting'' are appended to each prompt before generation. While our generation script includes a list of negative prompts (e.g., ``lowres, bad anatomy, blurry''), we note that Stable Diffusion 3.5 Medium often performs robustly without explicit negative prompting for common artifacts, and their impact in our specific workflow may be minimal.

\section{Experimental Setup}
\label{sec:experimental_setup}

To evaluate the efficacy of our pipeline in translating authorial styles into visual representations, we conducted a human evaluation study. This section details the dataset used, the design of our human evaluation, and the metrics employed for analysis.

\subsection{Dataset}
\label{ssec:dataset}
Our experiments leverage the Author Writing Sheets (AWS) for Reddit authors, as originally compiled and utilized in the Mythos dataset by \textit{Lan et al.} \cite{Kumar_StoryStyle_2025}. We processed the AWS for $N=49$ unique authors. For each author, our pipeline (detailed in Section~\ref{sec:methodology}) generated three distinct images, resulting in a total corpus of $49 \times 3 = 147$ images. Each set of three images is intended to offer varied visual interpretations of a single author's stylistic profile.

\subsection{Human Evaluation Design}
\label{ssec:human_eval_design}

\subsubsection{Participants}
\label{sssec:participants}
A cohort of $N=10$ individuals participated in the evaluation study. The participants comprised a diverse group with an age range of 21 to 48 years (mean age = 31.5, SD = 8.2). The gender distribution was 60\% male and 40\% female. Participants were recruited based on their general familiarity with creative content, digital media, and online platforms, ensuring an engaged audience capable of discerning stylistic nuances relevant to the study. No specific expertise in literary analysis or visual arts was required, aiming for feedback representative of a broader informed audience.

\subsubsection{Task}
\label{sssec:task}
Each of the 49 author profiles (consisting of an AWS and its corresponding three generated images) was independently evaluated by two different participants, resulting in a total of 98 author-set evaluations. To distribute the workload, each participant was assigned 10 unique ``Author Sets'' to evaluate (with one participant evaluating 8 sets to ensure complete coverage).
For each Author Set, participants were first presented with the cleaned AWS text summarizing the author's writing style across categories such as Plot, Creativity, Development, and Language Use. Subsequently, they were shown the three images (labeled A, B, and C) generated by our pipeline for that specific author.

\subsubsection{Survey Questions}
\label{sssec:survey_questions}
Participants responded to the following questions for each Author Set, designed to capture both quantitative and qualitative assessments:
\begin{itemize}
    \item \textbf{Q1 (Overall Style Match):} ``Considering the Author Writing Style Summary, how well do the three images (Image A, B, and C) \textbf{together} visually capture the author's overall aesthetic, mood, and characteristic themes?'' (Likert scale: 1-Not at all well to 5-Extremely well).
    \item \textbf{Q2 (Specific Stylistic Elements - Connection to Images):} ``From the Author Writing Style Summary, please identify \textbf{two specific stylistic elements} [...]. For each element: a) State the element. b) Explain how (or if) you see it reflected in \textbf{any} of the images (A, B, or C). [...]'' (Open-ended qualitative response).
    \item \textbf{Q3 (Best Individual Image Representation):} ``Which \textbf{one} of the three images (A, B, or C) do you feel \textbf{best} captures the author's overall style as described in the summary?'' (Choice: Image A, Image B, Image C, or None of them are a good fit). This was followed by a request to briefly explain their choice.
    \item \textbf{Q4 (Visual Distinctiveness):} ``How \textbf{distinctive} would you say the visual style is across these three images (considered as a set)? Do they feel like they represent a unique authorial vision, or are they more generic?'' (Likert scale: 1-Very Generic to 5-Exceptionally Distinctive).
\end{itemize}

\subsubsection{Data Collection}
\label{sssec:data_collection}
The survey was administered via a custom-built HTML interface. Each participant received a unique link that presented their assigned Author Sets sequentially. Responses were collected and compiled into a CSV file with columns: `id` (unique response ID), `rater-id`, `item-id` (author style sheet ID), `rating` (Q1 score), `favorite-image-id` (Q3 choice), and `distinctiveness` (Q4 score), along with qualitative responses.

\subsection{Evaluation Metrics}
\label{ssec:eval_metrics}
Our analysis relies on both quantitative and qualitative data derived from the survey.
\textbf{Quantitative Metrics:}
\begin{itemize}
    \item \textbf{Mean Overall Style Match:} Average score from Q1 for each author (averaged across two raters) and the overall mean across all authors.
    \item \textbf{Favorite Image Distribution:} Frequency distribution of choices for Q3 (Image A, B, C, or None) to identify if certain prompt variations were consistently preferred.
    \item \textbf{Mean Visual Distinctiveness:} Average score from Q4 for each author (averaged across two raters) and the overall mean.
    \item \textbf{Inter-Rater Reliability (IRR):} To assess consistency between raters, we will calculate:
        \begin{itemize}
            \item For `rating` (Q1) and `distinctiveness` (Q4): The percentage of evaluations where the two raters' scores were within $\pm 1$ point on the 5-point Likert scale, and the average absolute difference between scores.
            \item For `favorite-image-id` (Q3): The percentage of author styles for which both raters agreed on the single best representative image.
        \end{itemize}
\end{itemize}
\textbf{Qualitative Analysis:}
We will perform a thematic analysis of the open-ended responses from Q2 (stylistic element connections) and Q3 (justifications for favorite image). This will involve identifying recurring themes, illustrative examples of successful style translation, and common challenges or mismatches noted by the participants.

\section{Results}
\label{sec:results}

This section presents the findings from our human evaluation study, focusing first on the quantitative analysis of the collected ratings, followed by qualitative insights.

\subsection{Quantitative Findings}
\label{ssec:quantitative_findings}

\begin{figure*}[ht]
    \centering
    \begin{subfigure}[b]{0.48\textwidth}
        \centering
        \includegraphics[width=\textwidth]{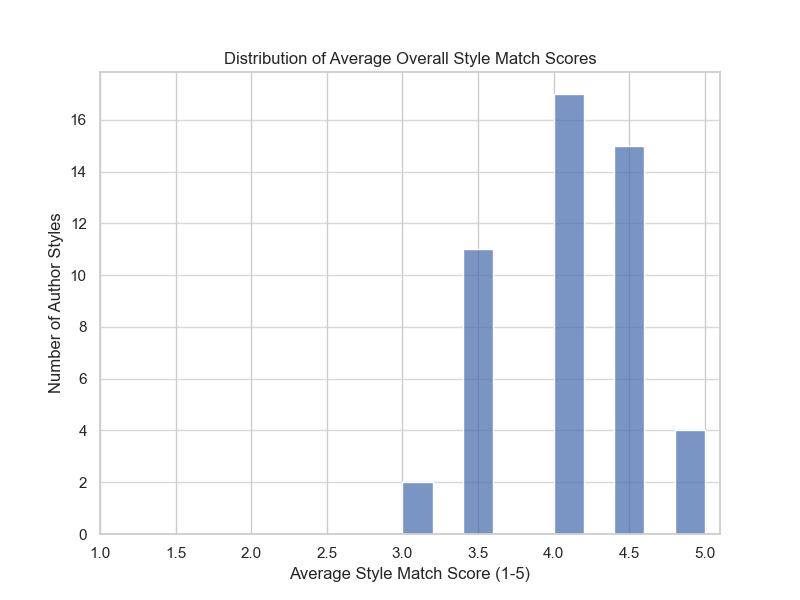}
        \caption{Distribution of average Overall Style Match scores (Q1) across the 49 author styles.}
        \label{fig:style_match_histogram_sub}
    \end{subfigure}
    \hfill
    \begin{subfigure}[b]{0.48\textwidth}
        \centering
        \includegraphics[width=\textwidth]{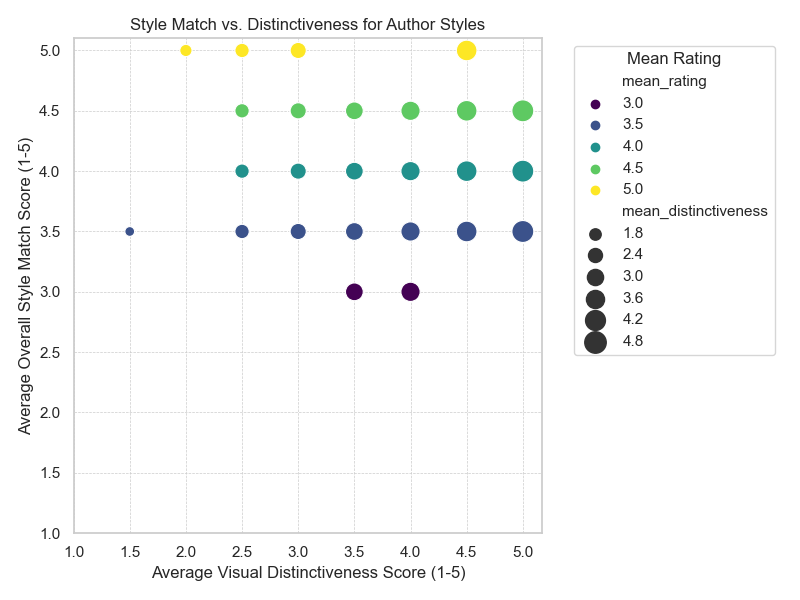}
        \caption{Scatter plot of average Overall Style Match (Q1) vs. average Visual Distinctiveness (Q4) scores.}
        \label{fig:distinctiveness_scatter_sub}
    \end{subfigure}
    \caption{Quantitative evaluation results: (a) shows the distribution of average Overall Style Match scores, indicating a generally positive reception. (b) plots average Style Match against average Visual Distinctiveness for each author style.}
    \label{fig:quantitative_visuals}
\end{figure*}

The quantitative data from our 10 evaluators, assessing 49 unique author styles (each with three generated images), provides insights into the perceived style match, image preference, visual distinctiveness, and inter-rater reliability. Each author style was evaluated by two different raters.

\subsubsection{Overall Style Match (Q1)}
\label{sssec:style_match}
Participants rated how well the set of three generated images visually captured the author's overall aesthetic, mood, and characteristic themes as described in the Author Writing Sheet (AWS), on a 5-point Likert scale (1=Not at all well, 5=Extremely well).
The mean of the per-item average ratings for Overall Style Match across all 49 author styles was $4.08$ (SD = $0.50$). This suggests that, on average, participants found the generated image sets to be a good representation of the authors' textual styles.

A histogram illustrating the distribution of these average per-item style match scores is presented in Figure~\ref{fig:quantitative_visuals}.

\subsubsection{Favorite Image Representation (Q3)}
\label{sssec:favorite_image}
Evaluators were asked to select which of the three images (Image 1, 2, or 3, corresponding to three distinct prompts) best captured the author's overall style.
Across all evaluations, Image 3 was chosen most frequently as the best representation (38.78\%), followed by Image 1 (33.67\%), and then Image 2 (27.55\%). This indicates a varied preference across the three prompt variations, with a slight inclination towards the third prompt's output.
For the 49 unique author styles, both raters agreed on the same single best representative image in 40.82\% of cases.

\subsubsection{Visual Distinctiveness (Q4)}
\label{sssec:distinctiveness}
The perceived visual distinctiveness of the image sets, whether they represented a unique authorial vision or were more generic, was rated on a 5-point Likert scale (1=Very Generic, 5=Exceptionally Distinctive).
The mean of the per-item average ratings for Visual Distinctiveness was $3.62$ (SD = $0.81$). This suggests a tendency towards perceiving the generated images as moderately distinctive.

Further quantitative insights are provided in Figure~\ref{fig:quantitative_visuals}, which illustrates the distribution of style match scores and the relationship between perceived style match and visual distinctiveness.

To assess the consistency of judgments between the two raters assigned to each author style, we calculated several IRR metrics.
For the \textbf{Overall Style Match (Q1)} ratings, the mean absolute difference between the two raters' scores was $0.735$ points on the 5-point scale. Raters provided scores within $\pm 1$ point of each other for 89.80\% of the author styles.
For \textbf{Visual Distinctiveness (Q4)} ratings, the mean absolute difference was $1.204$ points. Raters' scores were within $\pm 1$ point for 63.27\% of the author styles.
Regarding the \textbf{Favorite Image Representation (Q3)}, as previously noted, there was an agreement rate of 40.82\% on the specific image chosen as best representing the author's style.

These IRR scores indicate a higher consistency in judging the overall style match compared to visual distinctiveness or the specific best image, with style match ratings showing a strong tendency for raters to be in close agreement.

\subsection{Qualitative Insights}
\label{ssec:qualitative_insights}
In addition to the quantitative scores, qualitative feedback was collected to understand the nuances of how participants perceived the translation from textual authorial styles to visual representations. This feedback primarily stemmed from responses to Q2, where participants identified specific stylistic elements from the Author Writing Sheet (AWS) and described their manifestation (or lack thereof) in the generated images, and from Q3, which invited justifications for selecting the best representative image. While the level of detail in responses varied, the analysis of recurring themes and illustrative examples provides valuable context to the numerical data.

\subsubsection{Case Study: Visualizing Author ID \textit{Monsoon77}}
\label{sssec:case_study_1}
\begin{figure*}[ht]
    \centering
    \begin{subfigure}[b]{0.3\textwidth}
        \centering
        \includegraphics[width=\textwidth]{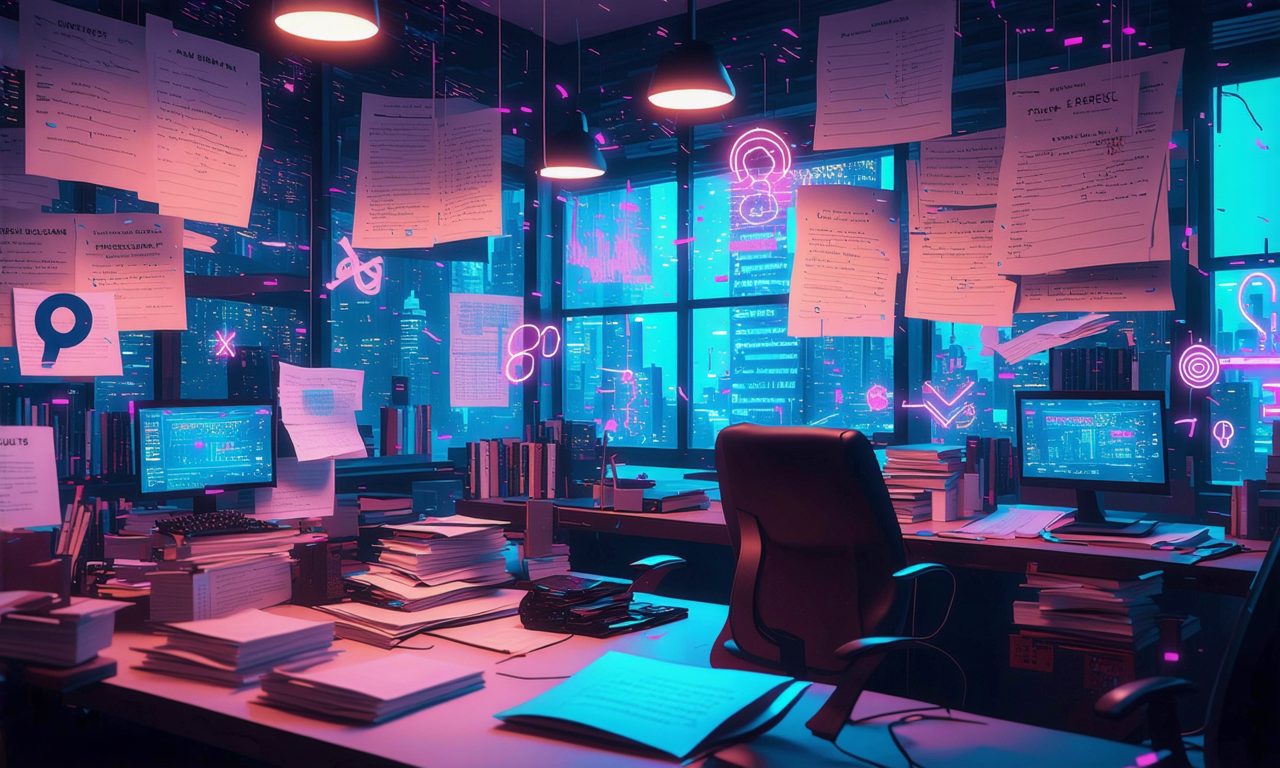} 
        \caption{Image A}
        \label{fig:case_study_img_a}
    \end{subfigure}
    \hfill
    \begin{subfigure}[b]{0.3\textwidth}
        \centering
        \includegraphics[width=\textwidth]{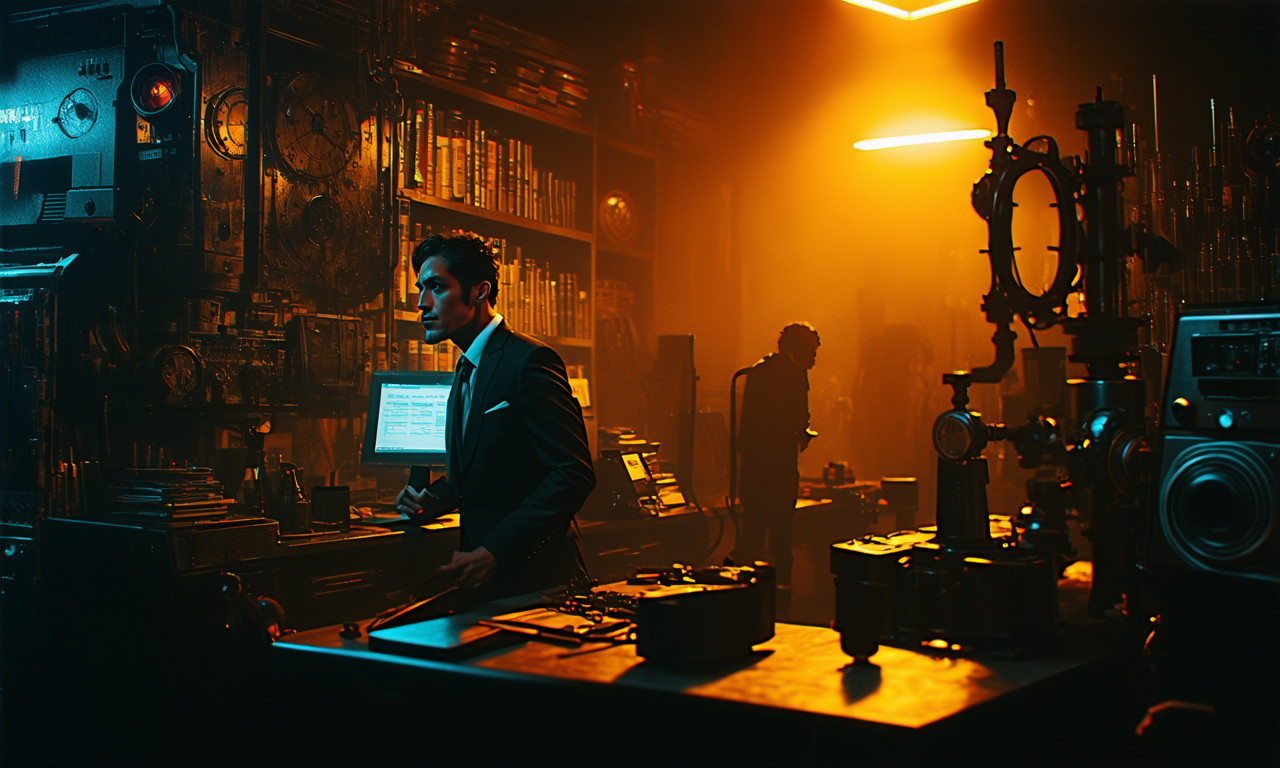} 
        \caption{Image B}
        \label{fig:case_study_img_b}
    \end{subfigure}
    \hfill
    \begin{subfigure}[b]{0.3\textwidth}
        \centering
        \includegraphics[width=\textwidth]{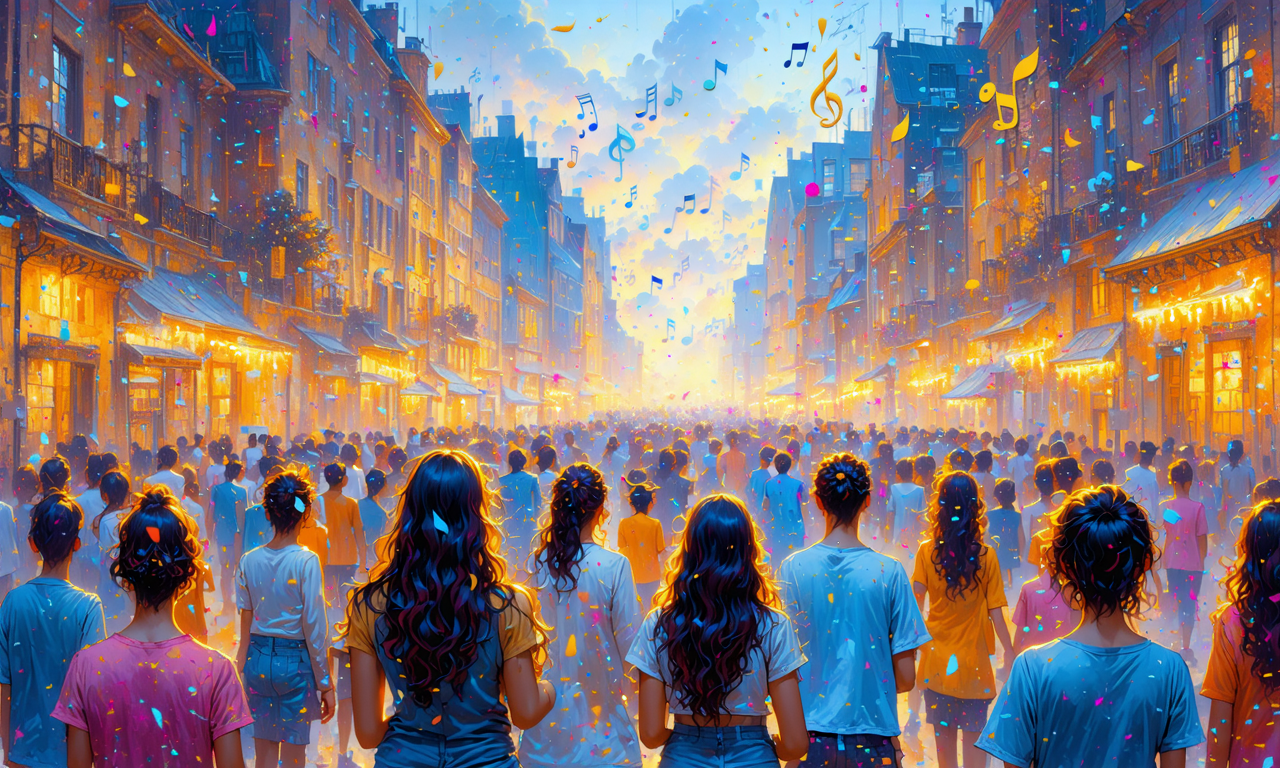} 
        \caption{Image C}
        \label{fig:case_study_img_c}
    \end{subfigure}
    \caption{Generated images for Author ID \textit{Monsoon77}, corresponding to the prompts derived from their AWS (see Table~\ref{tab:case_study_aws_prompts}).}
    \label{fig:case_study_images}
\end{figure*}

To illustrate the pipeline's capabilities and challenges, we examine the case of Author ID \textit{Monsoon77}. Table~\ref{tab:case_study_aws_prompts} presents key excerpts from this author's AWS and the corresponding core components of the image prompts generated by our LLM. The resulting images are shown in Figure~\ref{fig:case_study_images}.

\begin{table*}[ht]
\centering
\caption{AWS excerpts and corresponding generated image prompt components for Author ID \textit{Monsoon77}.}
\label{tab:case_study_aws_prompts}
\begin{tabular}{p{0.3\textwidth}|p{0.2\textwidth}|p{0.2\textwidth}|p{0.2\textwidth}}
\hline
\textbf{AWS Key Stylistic Claim} & \textbf{Image A Prompt Core} & \textbf{Image B Prompt Core} & \textbf{Image C Prompt Core} \\
\hline
\begin{itemize}
  \item \textbf{Plot Structure:} Stories vary from magical duels to internal conflicts, destiny resistance, moral debates, and suspenseful twists.
  \item \textbf{Creativity:} Authors subvert tropes, blend genres, and introduce unexpected elements to deepen themes and surprise readers.
  \item \textbf{Character and Setting:} Characters are revealed through dialogue, emotions, and social roles; settings reflect personal histories and values.
  \item \textbf{Language Use:} Vivid, sensory language and tailored dialogue create tension, emotion, and thematic depth.
\end{itemize}
&
\begin{flushleft}
magical office space, floating paperwork and glowing bureaucratic symbols, vibrant neon accents against corporate monotony, surreal perspective, dramatic lighting with supernatural glow, hyper-detailed 8k, absurdist fantasy illustration style, chaotic yet organized composition
\end{flushleft}
&
\begin{flushleft}
dimly lit vintage spy headquarters, exaggerated character silhouettes, humorous steampunk gadgets, cinematic film noir style, warm spotlighting, dynamic diagonal composition, highly detailed environment with retro-tech elements, 8k quality
\end{flushleft}
&
\begin{flushleft}
whimsical street celebration scene, diverse crowd with magical auras, soft twilight atmosphere, fairy lights and floating musical notes, dreamy bokeh effect, stylized watercolor illustration style, vibrant color palette, highly detailed 8k rendering
\end{flushleft}
\\
\hline
\end{tabular}
\end{table*}

Evaluator feedback for this author highlighted ``a strong capture of 'suspenseful twists and moral debates' in Image B''. One participant noted, ``Image C really nails the emotions and social roles. The way the musical notes, diverse crowd is portrayed, it highlights the thematic depth''
However, translating complex concepts like 'destiny resistance', 'surprise readers' appeared to be hard to capture in visual representation. Another evaluator commented on Image A, stating, ``While Image A has some interesting neon elements, it doesn't quite hit the "resistance or surprise" elements. It feels more like a standard moody sci-fi scene rather than something uniquely describing above mentioned properties.'' This suggests that while broad moods and settings might be effectively conveyed, highly specific or contrasting stylistic juxtapositions require further refinement in the prompt generation or image synthesis stages. Image C, chosen by one rater as the best fit, was praised for its depiction of `richly detailed', with the rater stating, ``The texture in the watercolored Image C perfectly matched the "vivid" and "social roles" aspect of the author's style sheet.''

\subsubsection{Recurring Themes in Qualitative Feedback}
\label{sssec:qual_themes}
Beyond individual case studies, several recurring themes emerged from the qualitative responses:
\begin{itemize}
    \item \textbf{Effective Mood and Atmosphere Translation:} Participants frequently acknowledged the pipeline's success in translating abstract mood descriptors from the AWS (e.g., ``somber'', ``playful'', ``mysterious'') into a palpable visual atmosphere in the generated images.
    \item \textbf{Challenge in Visualizing Complex/Abstract Narrative Elements:} A common observation was the difficulty in visually representing more complex narrative aspects, specific plot mechanics, or deeply internal character traits mentioned in the AWS. 
    \item \textbf{Varied Success with Specificity:} The translation of highly specific or unique stylistic elements  yielded mixed results, sometimes captured effectively and other times resulting in more generic interpretations, underscoring the sensitivity of the LLM interpretation and diffusion model rendering.
    \item \textbf{Impact of Prompt Nuance:} Some participants, particularly when justifying their choice for the 'best' image (Q3), indirectly highlighted the impact of different prompt phrasings, suggesting that certain visual interpretations (derived from the three distinct prompts) resonated more strongly with the AWS than others for the same author.
\end{itemize}
These qualitative insights underscore the promise of the proposed pipeline while also pinpointing areas for future refinement, particularly in enhancing the specificity and translatability of complex or highly abstract stylistic features.

\section{Discussion}
\label{sec:discussion}

Our findings suggest that the proposed pipeline, leveraging Author Writing Sheets (AWS) and LLM-driven prompt generation, represents a promising step towards visually personalizing authorial styles. The quantitative and qualitative results offer insights into the method's effectiveness, current limitations, and avenues for future exploration.

\subsection{Interpretation of Results}
\label{ssec:interpretation_results}
We now revisit our initial research questions in light of the experimental outcomes.

\textbf{RQ1 (Feasibility \& Effectiveness):} The mean Overall Style Match score of $4.08$ (out of 5) indicates that, on average, participants perceived a good alignment between the generated images and the authors' textual styles (as described in the AWS). This suggests that translating abstract literary characteristics into representative visual imagery is indeed feasible with our approach. The Inter-Rater Reliability for style match (89.80\% agreement within $\pm 1$ point) further lends confidence to this finding.

\textbf{RQ2 (Role of Intermediary LLM):} While not directly measured, the LLM's role as an interpreter between the AWS and image prompts is central. The generally positive style match scores imply that the LLM (Claude 3.7 Sonnet), guided by our prompting strategy, was largely successful in synthesizing visual descriptors from the textual style sheets. The varied success across different stylistic elements (discussed below) also points to the LLM's interpretation being a critical, and sometimes challenging, step. The preference for different image prompts (Image 3 being most favored at 38.78\%, but with Images 1 and 2 also frequently chosen) suggests that the LLM's generation of three distinct prompt variations was valuable, allowing for multiple facets of an author's style to be explored visually. This strategy likely captured different valid interpretations, as no single prompt variation universally outperformed others.

\textbf{RQ3 (Visual Personalization):} The mean Visual Distinctiveness score of $3.62$ (out of 5) suggests that the generated image sets were perceived as moderately distinctive, leaning away from being purely generic. While this is a positive indication of personalization, the lower IRR for distinctiveness (63.27\% agreement within $\pm 1$ point) compared to style match indicates greater subjectivity and potentially more rater variance in assessing uniqueness. The qualitative feedback, particularly when raters could connect specific AWS elements to visual outputs (as seen in Section~\ref{sssec:case_study_1}), further supports the idea that personalized visual traits were often discernible.

\textbf{RQ4 (Evaluation):} Our mixed-methods evaluation, combining quantitative ratings with qualitative justifications, proved valuable. The quantitative scores provided an overall measure of success, while the qualitative insights (Section~\ref{ssec:qualitative_insights}) were crucial for understanding \textit{why} certain translations worked and others were less effective.

The qualitative analysis highlighted that broader aspects like \textbf{mood and atmosphere} were often successfully translated, echoing findings from the original AWS paper \cite{Kumar_StoryStyle_2025} which noted that narrative categories like Creativity and Language Use (often tied to mood and tone) were easier to personalize textually than Plot. Conversely, similar to the challenges in textually personalizing specific plot structures, our visual pipeline found it difficult to convey \textbf{complex narrative elements or highly abstract internal character states} through static images. This suggests a shared challenge across modalities: the more abstract or narratively intricate the stylistic feature, the harder it is to capture and represent, whether in text or visuals.

\subsection{Strengths of the Approach}
\label{ssec:strengths}
Our work presents several notable strengths. Firstly, it introduces a \textbf{novel end-to-end pipeline} for a new task: visual authorial style personalization based on structured textual summaries. Secondly, the use of an LLM as a sophisticated interpreter to generate \textbf{stylistically-aware image prompts} from AWS offers a flexible and nuanced approach compared to manual prompt engineering for each author. Thirdly, the methodology's ability to generate \textbf{multiple distinct visual interpretations} (three images per author) from a single AWS allows for a richer exploration of an author's potential visual aesthetic.

\subsection{Limitations}
\label{ssec:limitations}
Despite promising results, this study has several limitations:
\begin{itemize}
    \item \textbf{Evaluation Scope:} The evaluation involved a limited sample of $N=10$ evaluators. Although each author style was rated twice, a larger and more diverse rater pool would enhance generalizability.
    \item \textbf{Subjectivity of Perception:} Assessing visual style and its consistency with a textual description is inherently subjective. While IRR metrics provide some measure of agreement, individual interpretations of both the AWS and the images inevitably vary.
    \item \textbf{Text-to-Image Model Capabilities:} Current text-to-image models, while powerful, still face challenges in accurately rendering highly complex, abstract, or subtly nuanced concepts described in textual prompts.
    \item \textbf{Data Source Specificity:} Our study was confined to AWS data from Reddit authors. Translating stylistic elements from other literary sources, including classic texts and varied genres, remains an area open for further study.
\end{itemize}

\subsection{Future Work}
\label{ssec:future_work}
This research opens several avenues for future investigation:
\begin{itemize}
    \item \textbf{Expanded Evaluation:} Conducting larger-scale human evaluations with demographically diverse and potentially expert raters (e.g., artists, literary critics) would provide more robust validation.
    \item \textbf{Model Exploration and Refinement:} Experimenting with different LLMs for prompt generation or more advanced text-to-image models, including those with enhanced stylistic control mechanisms, could yield improvements. Techniques for iterative refinement, where a user could guide or adjust the visual style interactively, also warrant exploration.
    \item \textbf{Diverse Data Sources:} Extending the methodology to AWS derived from other literary corpora (e.g., novels, poetry, screenplays) would test its versatility.
    \item \textbf{Automated Evaluation Metrics:} Investigating the potential of multimodal LLMs (VLMs) or other AI-driven techniques to develop automated or semi-automated metrics for cross-modal stylistic consistency could complement human evaluations.
    \item \textbf{Applications:} Exploring practical applications, such as generating inspirational mood boards for writers, concept art for characters or settings based on narrative style, or educational tools for visualizing literary techniques, would be valuable.
\end{itemize}

\section{Conclusion}
\label{sec:conclusion}

This paper introduced a novel pipeline for translating textually-defined authorial writing styles into visual representations. By leveraging Author Writing Sheets (AWS) as input to a Large Language Model (LLM) for descriptive prompt generation, and subsequently using a diffusion model for image synthesis, we demonstrated a promising approach to visual authorial style personalization. Human evaluation of 49 author styles indicated a good perceived alignment between the generated images and the textual AWS profiles (mean style match: $4.08/5$), with images being rated as moderately distinctive. Our key contributions include the end-to-end pipeline, the LLM-mediated methodology for generating stylistically-aware image prompts, and an initial empirical validation of this cross-modal translation.

While the visual translation of mood and atmosphere was often successful, representing highly abstract narrative elements remains an area for future work. This research pioneers a new direction in personalized generative AI, bridging literary style with visual art. Future efforts can focus on broader datasets, advanced model integration, and refined evaluation techniques, further unlocking the potential to visualize the unique essence of an author's narrative world for creative, educational, and entertainment applications.


\begin{thebibliography}{99} 

\bibitem{Kumar_StoryStyle_2025}
Kumar, Nischal Ashok, Chau Minh Pham, Mohit Iyyer, and Andrew Lan.
\newblock ``Whose Story Is It? Personalizing Story Generation by Inferring Author Styles.''
\newblock In \emph{arXiv preprint arXiv:2502.13028}, 2025.

\bibitem{wang_user_study_2024}
Wang, Ben, Jiqun Liu, Jamshed Karimnazarov, and Nicolas Thompson.
\newblock ``Task Supportive and Personalized Human-Large Language Model Interaction: A User Study.''
\newblock In \emph{Proceedings of the 2024 Conference on Human Information Interaction and Retrieval}, pp. 370--375, 2024.


\bibitem{zheng_feedback_llm_2023}
Zheng, Qichang, Tianjun Mo, and Xu Wang.
\newblock ``Personalized Feedback Generation Using LLMs: Enhancing Student Learning in STEM Education.''
\newblock \emph{Journal of Advanced Computing Systems}, vol. 3, no. 10, 2023, pp. 8--22.


\bibitem{chen_ai_theater_2024}
Chen, Nuo, Yan Wang, Yang Deng, and Jia Li.
\newblock ``The Oscars of AI Theater: A Survey on Role-Playing with Language Models.''
\newblock \emph{arXiv preprint arXiv:2407.11484}, 2024.


\bibitem{goodfellow_gan_2014_placeholder}
Goodfellow, Ian J., Jean Pouget-Abadie, Mehdi Mirza, Bing Xu, David Warde-Farley, Sherjil Ozair, Aaron Courville, and Yoshua Bengio.
\newblock ``Generative Adversarial Nets.''
\newblock In \emph{Advances in Neural Information Processing Systems 27 (NIPS 2014)}, pp. 2672--2680, 2014.

\bibitem{sohl_dickstein_diffusion_2015_placeholder}
Sohl-Dickstein, Jascha, Eric A. Weiss, Niru Maheswaranathan, and Surya Ganguli.
\newblock ``Deep Unsupervised Learning using Nonequilibrium Thermodynamics.''
\newblock In \emph{Proceedings of the 32nd International Conference on Machine Learning (ICML)}, pp. 2256--2265, 2015.

\bibitem{ho_denoising_diffusion_2020_placeholder}
Ho, Jonathan, Ajay Jain, and Pieter Abbeel.
\newblock ``Denoising Diffusion Probabilistic Models.''
\newblock In \emph{Advances in Neural Information Processing Systems 33 (NeurIPS 2020)}, pp. 6840--6851, 2020.

\bibitem{openai_dalle_2021_placeholder}
Ramesh, Aditya, Mikhail Pavlov, Gabriel Goh, Scott Gray, Chelsea Voss, Alec Radford, Mark Chen, and Ilya Sutskever.
\newblock ``Zero-Shot Text-to-Image Generation.''
\newblock In \emph{Proceedings of the 38th International Conference on Machine Learning (ICML)}, pp. 8821--8831, 2021.

\bibitem{saharia_imagen_2022_placeholder}
Saharia, Chitwan, William Chan, Saurabh Saxena, Lala Li, Jay Whang, Emily Denton, Seyed Kamyar Seyed Ghasemipour, Burcu Karagol Ayan, S. Sara Mahdavi, Rapha Gontijo Lopes, Tim Salimans, Jonathan Ho, David J. Fleet, and Mohammad Norouzi.
\newblock ``Photorealistic Text-to-Image Diffusion Models with Deep Language Understanding.''
\newblock In \emph{Advances in Neural Information Processing Systems 35 (NeurIPS 2022)}, pp. 36479--36493, 2022.

\bibitem{rombach_sd_2022_placeholder}
Rombach, Robin, Andreas Blattmann, Dominik Lorenz, Patrick Esser, and Björn Ommer.
\newblock ``High-Resolution Image Synthesis with Latent Diffusion Models.''
\newblock In \emph{Proceedings of the IEEE/CVF Conference on Computer Vision and Pattern Recognition (CVPR)}, pp. 10684--10695, 2022.

\bibitem{radford_clip_2021_placeholder}
Radford, Alec, Jong Wook Kim, Chris Hallacy, Aditya Ramesh, Gabriel Goh, Sandhini Agarwal, Girish Sastry, Amanda Askell, Pamela Mishkin, Jack Clark, Gretchen Krueger, and Ilya Sutskever.
\newblock ``Learning Transferable Visual Models From Natural Language Supervision.''
\newblock In \emph{Proceedings of the 38th International Conference on Machine Learning (ICML)}, pp. 8748--8763, 2021.

\bibitem{zhang_controlnet_2023}
Zhang, Lvmin, Anyi Rao, and Maneesh Agrawala.
\newblock ``Adding Conditional Control to Text-to-Image Diffusion Models.''
\newblock In \emph{Proceedings of the IEEE/CVF International Conference on Computer Vision (ICCV)}, pp. 3686--3697, 2023.

\bibitem{gatys_nst_2016_placeholder}
Gatys, Leon A., Alexander S. Ecker, and Matthias Bethge.
\newblock ``Image Style Transfer Using Convolutional Neural Networks.''
\newblock In \emph{Proceedings of the IEEE Conference on Computer Vision and Pattern Recognition (CVPR)}, pp. 2414--2423, 2016.

\bibitem{musiclm_placeholder_2023}
Agostinelli, Andrea, Timo I. Denk, Zalán Borsos, Jesse Engel, Mauro Verzetti, Antoine Caillon, Qingqing Huang, Aren Jansen, Adam Roberts, Marco Tagliasacchi, Matt Sharifi, Neil Zeghidour, and Christian Frank.
\newblock ``MusicLM: Generating Music From Text.''
\newblock \emph{arXiv preprint arXiv:2301.11325}, 2023.

\bibitem{Rombach_LatentDiffusion_2022}
Rombach, Robin, Andreas Blattmann, Dominik Lorenz, Patrick Esser, and Björn Ommer.
\newblock ``High-Resolution Image Synthesis with Latent Diffusion Models.''
\newblock In \emph{Proceedings of the IEEE/CVF Conference on Computer Vision and Pattern Recognition (CVPR)}, pp. 10684--10695, 2022.

\bibitem{Gal_TextualInversion_2022}
Gal, Rinon, Yuval Alaluf, Yuval Atzmon, Or Patashnik, Amit H. Bermano, Gal Chechik, and Daniel Cohen-Or.
\newblock ``An Image is Worth One Word: Personalizing Text-to-Image Generation using Textual Inversion.''
\newblock In \emph{International Conference on Learning Representations (ICLR)}, 2023. (Note: Often cited with its 2022 arXiv preprint, formal publication in ICLR 2023)

\bibitem{Ruiz_DreamBooth_2023}
Ruiz, Nataniel, Yuanzhen Li, Varun Jampani, Yael Pritch, Michael Rubinstein, and Kfir Aberman.
\newblock ``DreamBooth: Fine-Tuning Text-to-Image Diffusion Models for Subject-Driven Generation.''
\newblock In \emph{Proceedings of the IEEE/CVF Conference on Computer Vision and Pattern Recognition (CVPR)}, pp. 22500--22510, 2023.

\bibitem{zhang_personalization_survey_2024}
Zhang, Zhehao, Ryan A. Rossi, Branislav Kveton, Yijia Shao, Diyi Yang, Hamed Zamani, Franck Dernoncourt, et al.
\newblock ``Personalization of Large Language Models: A Survey.''
\newblock \emph{arXiv preprint arXiv:2411.00027}, 2024.


\bibitem{krubinski_headline_generation_2024}
Krubiński, Mateusz, and Pavel Pecina.
\newblock ``Towards Unified Uni- and Multi-Modal News Headline Generation.''
\newblock In \emph{Findings of the Association for Computational Linguistics: EACL 2024}, pp. 437--450, 2024.

\bibitem{li_review_generation_2019}
Li, Pan, and Alexander Tuzhilin.
\newblock ``Towards Controllable and Personalized Review Generation.''
\newblock \emph{arXiv preprint arXiv:1910.03506}, 2019.


\bibitem{chen_multimedia_tang_2024}
Chen, Xu, and Di Wu.
\newblock ``Automatic Generation of Multimedia Teaching Materials Based on Generative AI: Taking Tang Poetry as an Example.''
\newblock \emph{IEEE Transactions on Learning Technologies}, 2024.


\bibitem{li_prompt_learning_2023}
Li, Lei, Yongfeng Zhang, and Li Chen.
\newblock ``Personalized Prompt Learning for Explainable Recommendation.''
\newblock \emph{ACM Transactions on Information Systems}, vol. 41, no. 4, 2023, pp. 1--26.


\end{thebibliography}
\end{document}